\documentclass[dvipdfmx]{article}

\usepackage{amssymb}
\usepackage{latexsym}

\usepackage[dvipdfmx]{graphicx}

\usepackage{url}
\usepackage[dvipdfmx]{xcolor}
\definecolor{newcolor}{rgb}{.8,.349,.1}

\usepackage{amsmath}
\usepackage{amsthm}
\usepackage{cases}
\newtheorem{theorem}{Theorem}{}
{}
{}
{}

\begin{document}


















\title{Cancelable indexing based on low-rank approximation of correlation-invariant random filtering for fast and secure biometric identification}



\author{
Takao Murakami\thanks{National Institute of Advanced Industrial Science and Technology (AIST), Tokyo 135-0064, Japan}
\and 
Tetsushi Ohki\thanks{Shizuoka University, Hamamatsu 432-8011, Japan}
\and
Yosuke Kaga\thanks{Hitachi, Ltd., Yokohama 244-0817, Japan}
\and
Masakazu Fujio\thanks{Hitachi, Ltd., Yokohama 244-0817, Japan}
\and
Kenta Takahashi\thanks{Hitachi, Ltd., Yokohama 244-0817, Japan}
}


\maketitle

\begin{abstract}
A cancelable biometric scheme called correlation-invariant random filtering (CIRF) is known as a promising template protection scheme. 
This scheme transforms a biometric feature 
represented as an image 
via the 2D number theoretic transform (NTT) and random filtering. 
CIRF has perfect secrecy 
in that the transformed feature 
leaks no information 
about the original feature. 
However, CIRF cannot be applied to large-scale biometric identification, since the 2D inverse NTT in the matching phase requires high computational time. 
Furthermore, existing biometric indexing schemes cannot be used in conjunction with template protection schemes to speed up biometric identification, 
since a biometric index leaks some information about the original feature. 
In this paper, we propose 
a novel indexing scheme called ``cancelable indexing'' to speed up CIRF without losing its security properties. 
The proposed scheme is based on fast computation of CIRF via low-rank approximation of biometric images 
and via a minimum spanning tree representation of low-rank matrices in the Fourier domain. 
We prove that the transformed index 
leaks no information 
about the original index and the original biometric feature (i.e., perfect secrecy), 
and thoroughly discuss the security of the proposed scheme. 
We also 
demonstrate that 
it significantly reduces the one-to-many matching time using a finger-vein dataset that 
includes 
six fingers from 505 subjects.
\end{abstract}

{\bf{keyword}}:
cancelable biometrics,
biometric identification,
correlation-invariant random filtering,
indexing,
low-rank approximation,
minimum spanning tree.


\section{Introduction}
\label{sec:intro}

Biometric authentication systems, which recognize a person based on physical characteristics (e.g., fingerprint, finger-vein, iris) or behavioral characteristics (e.g., voice, gait), 
have been used for various applications (e.g., PC login, physical access control, banking). 
They are also expected to be applied to 
user authentication over networks (e.g., internet banking, online payment, membership authentication), 
where 
a client sends a biometric 
sample 
(referred to as a {\it query sample}) to a server for authentication, 
and the server compares it with a biometric feature enrolled in a database (referred to as a {\it template}). 

However, the use of biometric authentication over networks raises 
severe security concerns. 
Specifically, 
since biometric features (e.g., fingerprint, finger-vein, iris) are unchangeable, they cannot be revoked (unlike passwords or tokens) if they are leaked from the server. 
A naive way to prevent the leakage of biometric features is to encrypt the features using a conventional encryption scheme 
such as AES (Advanced Encryption Standard). 
However, since the encrypted features have to be decrypted to perform pattern matching on the server, a skilled attacker 
aiming 
at the timing of decryption can obtain the original features. 

Template protection schemes \cite{Jain_EURASIP08}, which keep biometric features secret even 
during 
the matching phase, have been widely studied to address this issue. 
They can be divided into two approaches: cancelable biometrics (a.k.a. feature transformation) 
\cite{Rathgeb_ICB13,biometric_security_ch1,Takahashi_IEICE11,Takahashi_IETBio12,Wong_PRL13} 
and biometric cryptosystems 
\cite{Dodis_JC08,Hartlogg_ICB13,Mai_BTAS15,Murakami_INFFUS16}. 
Among them, a cancelable biometric scheme called 
{\it correlation-invariant random filtering (CIRF)} \cite{biometric_security_ch1,Takahashi_IEICE11,Takahashi_IETBio12} is known as a promising template protection scheme. 
CIRF transforms 
biometric features 
via the 2D number theoretic transform (NTT) and random filtering. 
Then it performs pattern matching without restoring 
the original features 
by multiplying 
the transformed template by the transformed query sample 
and transforming 
the result 
via the 2D inverse NTT 
(see Section~\ref{sub:algorithm_CIRF} for details). 
CIRF can be applied to any 
kind of biometric traits whose score (distance or similarity) is measured 
via cross-correlation between biometric features 
(e.g., fingerprint \cite{Ishida_IEICE01}, face \cite{Brunelli_TPAMI93}, iris \cite{Daugman_TCSVT04}, finger-vein \cite{Miura_MVA04}) 
without affecting accuracy. 
In addition, 
it is proved in \cite{Takahashi_IEICE11,Takahashi_IETBio12} that CIRF has {\it perfect secrecy}. 
Specifically, 
according to \cite{Buchman_book}, a cryptosystem has perfect secrecy 
if any ciphertext $c \in \mathbf{C}$ ($\mathbf{C}$: ciphertext space) provides 
no information about the plaintext $m \in \mathbf{P}$ ($\mathbf{P}$: plaintext space); 
i.e., $\Pr(m|c) = \Pr(m)$ for any $m \in \mathbf{P}$ and any $c \in \mathbf{C}$. 
Similarly, it is proved in \cite{Takahashi_IEICE11,Takahashi_IETBio12} that in CIRF, 
the transformed template $T \in \mathbf{T}$ 
($\mathbf{T}$: space of transformed templates) 
provides no information about 
the original template $X \in \mathbf{X}$ 
($\mathbf{X}$: space of original templates); 
i.e., $\Pr(X|T) = \Pr(X)$ for any $X \in \mathbf{X}$ and any $T \in \mathbf{T}$. 

However, 
CIRF 
cannot be applied to a large-scale biometric identification system 
because of 
its high computational time during the matching phase. 
More specifically, 
biometric authentication can operate 
in either of 
the following two modes: verification and identification \cite{guide}. 
In the verification mode, 
a user claims an identity 
(i.e., enters an ID number or presents a smart card)
and inputs his/her query sample. 
Then 
the system compares the query sample with a template corresponding to the claimed identity 
(i.e., one-to-one matching).  
In the identification mode, 
a user 
inputs only his/her query sample. 
Then 
the system compares the query sample 
with many templates in the database (i.e., one-to-many matching). 
Based on the 
scores, the system identifies the user 
(e.g., if the system finds a template whose 
distance falls below a threshold, it identifies the user as the corresponding enrollee; 
if there is no such template, it rejects the user). 
Biometric identification offers a more convenient way of authentication, 
since a user 
need not 
enter an ID number 
or 
present a smart card. 
However, the response time can be very long in large-scale biometric identification, 
since the one-to-many matching time increases in proportion to the number of templates. 
This problem is particularly severe for CIRF, 
since the 2D inverse NTT in the matching phase requires 
a number of 1D inverse NTTs, which involve high computational time. 
For example, in our experiments in Section~\ref{sec:exp}, it took $0.28$ ms to compute a score between two biometric features based on CIRF. 
This means that it takes 
about $9$ ($\approx 0.28 \times 10^{-3} \times 32000$) 
seconds to identify a user when the number of templates in the database is $32000$. 

A biometric indexing (or classification) scheme 
\cite{encyclopedia,class_index} 
has been widely studied to speed up biometric identification 
(see Section~\ref{sub:related} for details). 
It computes, for each biometric feature, an {\it index}, which is a simple representation of 
the biometric feature 
(e.g., binary string, vector). 
In the identification phase, 
it typically computes an {\it approximate score (distance or similarity)} for each template using the index. 
In this paper, we refer to this process as an {\it approximate matching}. 
After the approximate matching, it sorts templates in ascending (or descending) order of the approximate distance (or similarity) and 
computes an ``exact'' score for a template according to the sorted order 
(e.g., if the exact distance falls below a threshold, identify the user as the corresponding enrollee; otherwise, continue to the next template). 
Since the index is a simple representation of the biometric feature, 
the approximate score 
is efficiently computed and 
is highly correlated with an exact score. 
Therefore, a genuine template can be found in the early stage of exact matching. 
Consequently, 
the number of exact score computations can be significantly reduced. 

It should be noted, however, that a biometric index 
leaks 
some information about the original biometric feature, 
since it is a simple representation of a biometric feature. 
Therefore, a biometric index 
needs to be protected in the same way as 
a biometric feature when it is used to speed up biometric identification over networks. 
In the following, 
we review previous work related to this issue. 

\subsection{Related Work}
\label{sub:related}

Biometric indexing schemes have been widely studied in the literature (a survey of them can be found in \cite{encyclopedia,class_index}). 
They are recently studied for 
various kinds of biometric traits; e.g., 
fingerprint \cite{Cappeli_TPAMI11,Shuai_ICPR08,Wang_TIFS15}, 
iris \cite{Dey_TIFS12,Proenca_TIFS13,Rathgeb_ICB15}, and 
finger-vein \cite{Kavati_SSCC13,Raghavendra_ISBA15,Tang_ICPR10}. 
However, 
most existing biometric indexing 
schemes 
do not protect a biometric index, 
and cannot be 
securely 
used for 
biometric identification over networks. 
Some studies \cite{Hartlogg_ICB13,Jin_TIFS18,Kuzu_ICDE12,Wang_ICDE13,Yiu_TKDE12} proposed 
an indexing (or hashing)  scheme that transforms or encrypts 
an index (or hash) and performs a 
query search without recovering the original index (or hash). 
However, 
these schemes 
do not guarantee that the transformed index 
leaks no information 
about the original index 
(i.e., perfect secrecy). 
As a different approach, a filtering technique for biometric identification based on secure multiparty computation 
was 
proposed in \cite{Bringer_ICB12}. 
In this 
approach, 
however, the original templates are stored in the server, 
and can be leaked by illegal access or 
by 
internal fraud.

\subsection{Our contributions}
\label{sub:contribution}
In this paper, we propose 
a novel indexing scheme called {\it cancelable indexing} to speed up CIRF without losing its security properties. 
The proposed indexing scheme has 
perfect secrecy in that the transformed index 
leaks no information 
about the original biometric feature. 
To our knowledge, the proposed indexing scheme 
is the first to have 
such perfect secrecy. 
Our contributions are 
as follows: 
\begin{itemize}
\item We propose a {\it cancelable indexing} scheme 
based on fast computation of CIRF via low-rank approximation of biometric images 
and via a minimum spanning tree representation of low-rank matrices in the Fourier domain. 
We prove that the proposed indexing scheme computes 
a cross-correlation between two approximated biometric images with much less 1D inverse NTTs 
(Section~\ref{sub:analysis}, \textbf{Theorem 1}). 
\item We then prove that the transformed index 
leaks no information 
about the original biometric feature and the original index 
(Section~\ref{sub:analysis}, \textbf{Theorem 2}). 
To our knowledge, the proposed indexing scheme is the first to have such perfect secrecy. 
Based on this 
property, we thoroughly discuss the security of the proposed scheme. 
\item We evaluate the proposed scheme using the finger-vein dataset in \cite{Yanagawa_BIC07}, which 
includes 
six fingers from $505$ subjects. 
Our experimental 
results show that the proposed scheme 
significantly reduces the one-to-many matching time. 
For example, the proposed scheme requires only about one second on average when the number of 
templates is $32000$ (whereas 
it takes about $9$ seconds to compute exact scores for all of the templates). 
\end{itemize}



\section{Preliminaries}
\label{sec:CIRF}
In this section, we 
describe details of 
cancelable biometrics and 
CIRF 
\cite{biometric_security_ch1,Takahashi_IEICE11,Takahashi_IETBio12}. 
We first explain an overview of cancelable biometrics in the case of verification 
and identification in Sections~\ref{sub:cancelable} and \ref{sub:cancelable_identification}, respectively. 
We then describe desirable properties for cancelable biometrics in Section~\ref{sub:desirable_cancelable}. 
We finally explain 
details of CIRF 
in Section~\ref{sub:algorithm_CIRF}. 

\subsection{Cancelable biometrics for verification}
\label{sub:cancelable}

\begin{figure}
\begin{center}
 \includegraphics[width=0.88\linewidth]{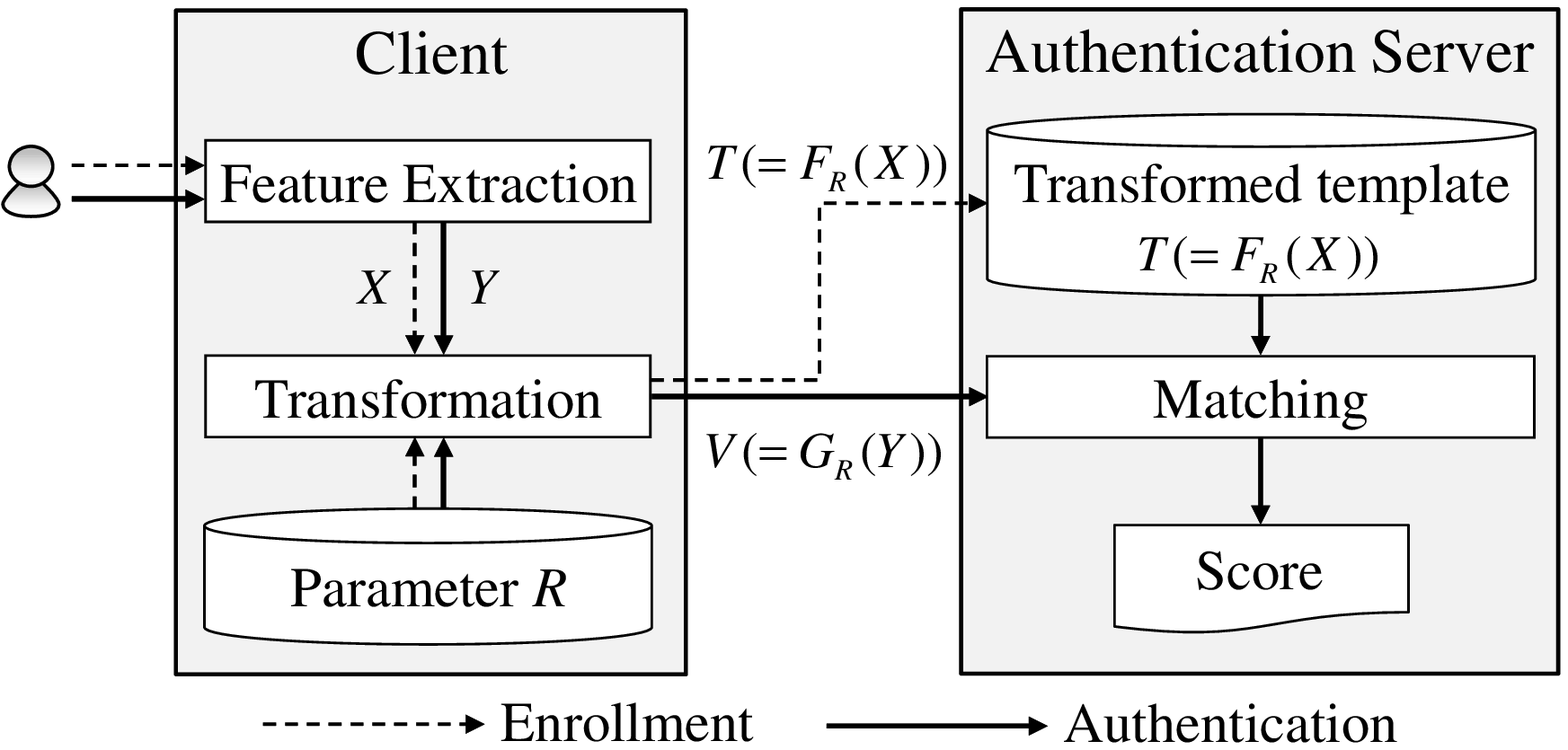}
 \caption{Overview of cancelable biometrics ($X$: template, $Y$: query sample, $R$: parameter, $T$: transformed template, $V$: transformed query sample). 
}
 \label{fig:cancelable}
\end{center}
\end{figure}

Fig.~\ref{fig:cancelable} shows 
an overview of 
cancelable biometrics in the case of biometric verification. 
Let $\mathbf{X}$, $\mathbf{Y}$, $\mathbf{R}$, $\mathbf{T}$, and $\mathbf{V}$ be 
spaces of 
templates, query samples, parameters, transformed templates, and transformed query samples, respectively. 
We denote a template, query sample, parameter, transformed template, and transformed query sample 
by $X \in \mathbf{X}$, $Y \in \mathbf{Y}$, $R \in \mathbf{R}$, $T \in \mathbf{T}$, and $V \in \mathbf{V}$, respectively. 
In the enrollment phase, a template $X$ is transformed via 
a transformation function $F_R: \mathbf{X} \rightarrow \mathbf{T}$, 
which is dependent on a parameter $R$, 
and the transformed template $T = F_R(X)$ is stored in an authentication server. 
The parameter $R$ is 
uniformly randomly generated, 
and plays 
a role similar to 
an encryption key. 
The parameter $R$ can be stored in a client or a {\it parameter management server} \cite{Takahashi_CIBIM11}, which is administered separately from the authentication server. 

In the authentication phase, a query sample $Y$ is transformed via a transformation function $G_R: \mathbf{Y} \rightarrow \mathbf{V}$, 
which is dependent on the parameter $R$, 
and the transformed 
query sample 
$V = G_R(Y)$ is sent to the authentication server. 
The authentication server compares 
$V$ ($= G_R(Y)$) 
with the transformed template $T$ ($= F_R(X)$). 
Let $\mathbb{R}$ be the set of real numbers, and 
$s: \mathbf{X} \times \mathbf{Y} \rightarrow \mathbb{R}$ be an 
\textit{exact score function} that takes input a template $X$ and a query sample $Y$ 
and outputs an exact score (distance or similarity) 
$s(X,Y) \in \mathbb{R}$ between $X$ and $Y$. 
The authentication server computes an exact score $s(X,Y)$ 
by comparing $V$ with $T$ \textit{without restoring $X$ and $Y$}. 
If the distance (resp.~similarity) $s(X,Y)$ is smaller (resp.~larger) than a pre-determined threshold, the authentication server accepts the user (otherwise, it rejects the user). 
The transformation function $F_R$ is designed so that 
the original biometric feature $X$ cannot be recovered from 
the transformed template $F_R(X)$. 
Therefore, even if 
$F_R(X)$ or 
$R$ is leaked, 
they can be revoked by generating a new parameter $R_{new}$ and 
replacing $F_R(X)$ with a new transformed template $F_{R_{new}}(X)$. 
Similarly, $G_R$ is designed so that $Y$ cannot be recovered from $G_R(Y)$. 

Note that 
the original template $X$ can be recovered from $F_R(X)$ and $R$, 
if $F_R$ is a bijective (one-to-one) function. 
Even if $F_R$ is a many-to-one function, 
it is possible to recover $X$ from $F_R(X)$ and $R$, 
as shown in \cite{Nagar_MFS10,Quan_ISCSCT08}. 
Therefore, if both $F_R(X)$ and $R$ are leaked, $X$ can be recovered from them. 
It is important to note, however, that 
if the parameter $R$ is managed separately from 
the transformed template $F_R(X)$ 
(e.g., $R$ is stored in 
the client or the parameter management server \cite{Takahashi_CIBIM11}), 
the risk of simultaneous leakage of $F_R(X)$ and $R$ can be significantly reduced. 
If we store $R$ in a client that can be accessed by any user 
(e.g., ATM, POS, kiosk terminal), 
there might be a high risk that $R$ is leaked from the client. 
Thus, a more secure way would be to use a parameter management server \cite{Takahashi_CIBIM11}. 
In this model, 
a 
parameter management server securely manages $R$. 
The parameter management server and the authentication server 
are administered separately by different administrators or organizations, 
and they do not collude with each other. 
Since users cannot access the parameter management server, 
the risk of the leakage of $R$ is much smaller 
(for details of the authentication protocol in this model, see \cite{Takahashi_CIBIM11}). 

Separate and secure management and periodic revocation of $F_R(X)$ and $R$ play 
important roles 
in protecting templates. 

\subsection{Cancelable biometrics for identification}
\label{sub:cancelable_identification}

We now explain cancelable biometrics in the case of biometric identification. 
Let $\mathbb{N}$ be the set of natural numbers, 
$N \in \mathbb{N}$ be the number of templates, 
$X^{(n)} \in \mathbf{X}$ be the $n$-th template, 
and $T^{(n)} \in \mathbf{T}$ be the $n$-th transformed template ($1 \leq n \leq N$). 
Let further $R^{(n)} \in \mathbf{R}$ be a parameter for the $n$-th template $X^{(n)}$. 
Regarding 
parameters $R^{(1)}, \cdots, R^{(N)}$, we consider the following two scenarios: 
(i) $R^{(1)}, \cdots, R^{(N)}$ are independently and uniformly generated 
(i.e., a parameter is different from template to template); 
(ii) a common parameter $R^*$ ($= R^{(1)} = \cdots = R^{(N)}$), which is uniformly generated, 
is used for all templates.  
We refer to the former scenario as an {\it individual parameter scenario}, 
and the latter as a {\it common parameter scenario}. 

In the individual parameter scenario, 
the $n$-th template $X^{(n)}$ is transformed using the $n$-th parameter $R^{(n)}$: 
$T^{(n)} = F_{R^{(n)}}(X^{(n)})$ ($1 \leq n \leq N$). 
If we store all parameters $R^{(1)}, \cdots, R^{(N)}$ in the client, 
the client transforms a query sample $Y$ using each of $R^{(1)}, \cdots, R^{(N)}$ in the authentication phase. 
Let $V^{(n)} = G_{R^{(n)}}(Y) \in \mathbf{V}$ be the $n$-th transformed query sample. 
The client sends $N$ transformed query samples $V^{(1)}, \cdots, V^{(N)}$ to the authentication server. 
Then the authentication server computes a score between 
$V^{(n)}$ (= $G_{R^{(n)}}(Y)$) 
and $T^{(n)}$ ($= F_{R^{(n)}}(X^{(n)})$) ($1 \leq n \leq N$). 
In this case, 
the communication cost between the client and the authentication server is 
proportional to $N$. 
Similarly, if we store $R^{(1)}, \cdots, R^{(N)}$ 
in the parameter management server \cite{Takahashi_CIBIM11}, 
the communication cost between the parameter management server and the authentication 
server is proportional to $N$. 

The common parameter scenario is much more efficient in terms of the communication cost. 
In this case, the $n$-th template $X^{(n)}$ is transformed using 
a common parameter $R^*$: 
$T^{(n)} = F_{R^*}(X^{(n)})$ ($1 \leq n \leq N$). 
At the authentication phase, 
we only have to transform a query sample $Y$ using $R^*$ and 
send the transformed query sample $V^* = G_{R^*}(Y) \in \mathbf{V}$ 
to the authentication server. 
Then the authentication server 
computes a score between $V^*$ ($= G_{R^*}(Y)$) 
and $T^{(n)}$ ($= F_{R^*}(X^{(n)})$). 
In this case, 
the communication cost does not depend on $N$. 
However, since we use the same parameter $R^*$ for all templates in this case, 
we need to thoroughly discuss its security. 
In this paper, we propose a cancelable indexing scheme for 
both the individual parameter scenario and the common parameter scenario, 
and thoroughly discuss the communication cost and the security in both cases in 
Section~\ref{sub:analysis}.

\subsection{Desirable properties for cancelable biometrics}
\label{sub:desirable_cancelable}
An ideal cancelable biometric system should have the following properties \cite{ISO24745,Jain_EURASIP08}: 
\begin{enumerate}
\renewcommand{\labelenumi}{(\roman{enumi})}
\item \textbf{Security (irreversibility):} It should be impossible or computationally hard to recover the original biometric feature from the transformed feature. 
\item \textbf{Diversity (unlinkability):} The transformed feature should not allow cross-matching 
across databases. 
\item \textbf{Revocability:} It should be straightforward to revoke a compromised template and reissue a new one based on the same biometric data. 
\item \textbf{Accuracy:} FAR (False Acceptance Rate) and FRR (False Rejection Rate) should not be degraded by transforming biometric features. 
\end{enumerate}
In biometric identification, an ideal cancelable biometric system should also have the following property: 
\begin{enumerate}
\renewcommand{\labelenumi}{(\roman{enumi})}
\setcounter{enumi}{4}
\item \textbf{Response time:} The one-to-many matching time should be small (e.g., one second). 
\end{enumerate}

Regarding the security (irreversibility) in biometric identification, we consider 
attackers of the following three types: 
\begin{itemize}
\item \textbf{Attacker A} who obtains one transformed template, which 
corresponds 
to the $n$-th template $X^{(n)}$. 
\item \textbf{Attacker B} who obtains $N$ transformed templates, which 
corresponds 
to the templates $X^{(1)}, \cdots, X^{(N)}$. 
\item \textbf{Attacker C} who obtains all transformed features, which 
corresponds 
to the templates $X^{(1)}, \cdots, X^{(N)}$ and the query sample $Y$. 
\end{itemize}
Typically, \textbf{Attackers A} and \textbf{B} are outsiders who obtain the transformed template(s) leaked from the authentication server (i.e., external attackers), 
whereas 
\textbf{Attacker C} is a malicious server (i.e., internal attacker). 
For 
the 
cancelable indexing scheme proposed in Section~\ref{sec:cancelable_index}, 
we assume that 
these attackers obtain transformed indexes in addition to transformed features. 

\subsection{Correlation-invariant Random Filtering (CIRF)}
\label{sub:algorithm_CIRF}

We describe an algorithm for CIRF \cite{biometric_security_ch1,Takahashi_IEICE11,Takahashi_IETBio12} in biometric verification (we can extend it to biometric identification, as described in Section~\ref{sub:cancelable_identification}). 

\begin{figure}
\begin{center}
 \includegraphics[width=0.8\linewidth]{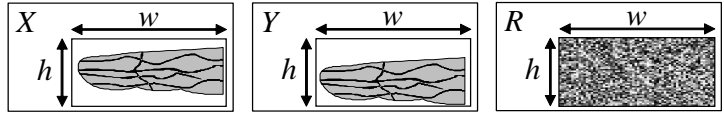}
 \caption{Template $X$, query sample $Y$, and parameter $R$ in CIRF.}
 \label{fig:fingervein}
\end{center}
\end{figure}

We assume that a biometric feature is represented as an image with 
$h$ (height) $\times$ $w$ (width) pixels (as shown in Fig.~\ref{fig:fingervein}), 
and each pixel value is a nonnegative integer 
less than $p$ ($p$ is a sufficiently large prime, which is described later in detail). 
Let $\mathbb{Z}_p$ be the set of nonnegative integers less than $p$; 
i.e., $\mathbb{Z}_p = \{0, 1, \cdots, p-1\}$. 
Then, a template and a query sample can be expressed as 
$X \in \mathbb{Z}_p^{h \times w}$ and $Y \in \mathbb{Z}_p^{h \times w}$, 
respectively; i.e., $\mathbf{X} = \mathbf{Y} = \mathbb{Z}_p^{h \times w}$. 
Let 
$X[i,j]$ and $Y[i,j]$ 
($0 \leq i < h, 0 \leq j < w$) be 
the $(i,j)$-th pixel of $X$ and $Y$, respectively.  
We assume that 
an exact score $s(X,Y) \in \mathbb{R}$ 
between $X$ and $Y$ 
is computed using 
cyclic cross-correlation $X \star Y$: 
\begin{align}
& (X \star Y)[\Delta i, \Delta j] 
= \sum_{i=0}^{h-1} \sum_{j=0}^{w-1} X[i,j] Y[i+\Delta i~\text{mod}~h, j+\Delta j~\text{mod}~w] \nonumber \\
& (- \Delta i_{max} \leq \Delta i \leq \Delta i_{max}, - \Delta j_{max} \leq \Delta j \leq \Delta j_{max} ), 
\label{eq:x_star_y}
\end{align}
where $\Delta i_{max}$ $(< h)$ and $\Delta j_{max}$ $(< w)$ are maximum allowable shift lengths between $X$ and $Y$. 
$X \star Y$ can also be expressed as cyclic convolution $X \ast \overleftarrow{Y}$, where $\overleftarrow{Y}$ denotes a flipped image of $Y$: $\overleftarrow{Y}[i,j] = Y[h-i-1, w-j-1]$. 
Since linear cross-correlation can be computed using cyclic cross-correlation with the 
help 
of zero-padding, 
this assumption is valid for any kind of 
biometric traits whose score $s(X,Y)$ is 
measured via 
linear cross-correlation \cite{Brunelli_TPAMI93,Daugman_TCSVT04,Ishida_IEICE01,Miura_MVA04}. 
Examples of such scores include the minimum of the Hamming distances of overlapped binary images 
over all values of $\Delta i$ and $\Delta j$ 
(see \ref{app:Hamming} for details). 

CIRF transforms a template $X$ via the 2D number theoretic transform (NTT) 
$\mathcal{F}: \mathbb{Z}_p^{h \times w} \rightarrow \mathbb{Z}_p^{h \times w}$, 
a kind of 2D discrete Fourier transform (DFT) over the Galois field $\mathbb{Z}_p$: 
\begin{align}
\mathcal{F}(X)[u,v] = 
\sum_{i=0}^{h-1} \sum_{j=0}^{w-1} \alpha^{ui} \beta^{vj} X[i,j]~\text{mod}~p. 
\label{eq:F_X_u_v}
\end{align}
$\alpha, \beta \in \mathbb{Z}_p$ are elements of the Galois field $\mathbb{Z}_p$ whose orders are $h$ and $w$, respectively. 
$p$ is a prime that satisfies 
\begin{align}
h ~|~ p -1~~\text{and}~~w ~|~ p -1
\end{align}
(i.e., $h$ and $w$ divide $p-1$), and is larger than the maximum of the cyclic cross-correlation between two biometric features \cite{Takahashi_IEICE11}. 
Hereinafter, we assume that all numerical operations are performed over $\mathbb{Z}_p$, and omit the notation ``$\text{mod}~p$''. 

After computing $\mathcal{F}(X)$, CIRF computes $\mathcal{F}(X) \circ R$, 
where $\circ$ denotes a pixel-wise multiplication (i.e., Hadamard product) 
and $R$ is a parameter (random filter) uniformly randomly generated from $(\mathbb{Z}_p^*)^{h \times w}$ ($\mathbb{Z}_p^* = \mathbb{Z}_p \setminus \{0\}$). 
In other words, the space of parameters is 
$\mathbf{R} = (\mathbb{Z}_p^*)^{h \times w}$. 
As for 
the 
query sample $Y$, CIRF 
computes $\mathcal{F}(\overleftarrow{Y}) \circ R^{-1}$, 
where $R^{-1}$ 
denotes 
a pixel-wise inverse of $R$. 
Thus, the transformed template $T$ and the transformed query sample $V$ are 
$T = F_R(X) = \mathcal{F}(X) \circ R$ and $V = G_R(Y) = \mathcal{F}(\overleftarrow{Y}) \circ R^{-1}$, 
respectively, and $\mathbf{T} = \mathbf{V} = \mathbb{Z}_p^{h \times w}$. 

In the matching phase, CIRF computes $\mathcal{F}^{-1}(T \circ V)$ ($\mathcal{F}^{-1}$ is the 2D inverse NTT), which can be written as follows: 
\begin{align}
\mathcal{F}^{-1}(T \circ V) 
&= \mathcal{F}^{-1}((\mathcal{F}(X) \circ R) \circ (\mathcal{F}(\overleftarrow{Y}) \circ R^{-1})) \\
&= \mathcal{F}^{-1}(\mathcal{F}(X) \circ \mathcal{F}(\overleftarrow{Y})) = X \ast \overleftarrow{Y} = X \star Y \label{eq:x_ast_y}
\end{align}
In other words, 
CIRF computes 
cross-correlation $X \star Y$ without restoring 
the original biometric features $X$ and $Y$. 
Therefore, CIRF can be applied to any kind of 
biometric traits whose score is measured 
via $X \star Y$ without affecting accuracy.

It is proved that the transformed template $T$ 
leaks no information 
about the original template $X$: 
$\Pr(X|T) = \Pr(X)$ for any $X \in \mathbb{Z}_p^{h \times w}$ and any $T \in \mathbb{Z}_p^{h \times w}$ (i.e., perfect secrecy) \cite{Takahashi_IEICE11,Takahashi_IETBio12}. 
Similarly, $V$ 
leaks no information 
about $Y$: $\Pr(Y | V) = \Pr(Y)$ for any $Y \in \mathbb{Z}_p^{h \times w}$ and any $V \in \mathbb{Z}_p^{h \times w}$. 
If the transformed template $T$ or the parameter $R$ is leaked, 
they can be revoked 
by generating a new parameter $R_{new}$ and issuing a new transformed template $T_{new}$ as follows: 
$T_{new} = T \circ (R_{new} \circ R^{-1})$ \cite{Takahashi_IEICE11}.  

In addition, it is proved that two transformed templates $T_1 = F_{R_1}(X)$ and $T_2 = F_{R_2}(X)$ generated from 
the same biometric feature $X$ are independent: 
$\Pr(T_1|T_2) = \Pr(T_1)$ if $R_1$ and $R_2$ are independently and uniformly generated \cite{Takahashi_IEICE11}. 
Therefore, the attacker cannot perform cross-matching across databases. 

In summary, CIRF is a promising scheme with regard to the properties (i), (ii), (iii), and (iv) in Section~\ref{sub:desirable_cancelable}. 

\section{Cancelable indexing based on Low-rank Approximation of CIRF}
\label{sec:cancelable_index}

A major 
shortcoming 
of CIRF is 
that it cannot be applied to a large-scale biometric identification 
because of 
high computational cost in the matching phase 
(i.e., it does not have the property (v) in Section~\ref{sub:desirable_cancelable}). 
In particular, the 2D inverse NTT of $T \circ V$ 
requires high computational cost (we confirmed that the time to compute the 2D inverse NTT accounts for most of the matching time). 
The 2D inverse NTT of $T \circ V$ requires 
$h + w$ separate 1D inverse NTTs, even with the aid of the row-column algorithm. 
Therefore, 
we propose a cancelable indexing scheme that significantly reduces the number of the 1D inverse NTTs. 

We first 
explain an overview of the proposed indexing scheme in Section~\ref{sub:overview}. 
We then 
describe an algorithm for the proposed indexing scheme in Sections~\ref{sub:algorithm} and \ref{sub:computation_M}. 
We 
finally 
explain its theoretical properties in Section~\ref{sub:analysis}. 

\subsection{Overview}
\label{sub:overview}

We begin by briefly explaining an overview of the proposed indexing scheme. 
Let 
$N \in \mathbb{N}$ be the number of transformed template stored in an authentication server, 
$X^{(n)} \in \mathbb{Z}_p^{h \times w}$ $(1 \leq n \leq N)$ be the $n$-th template, 
and $Y \in \mathbb{Z}_p^{h \times w}$ be a query sample. 

\begin{figure}
\begin{center}
 \includegraphics[width=0.7\linewidth]{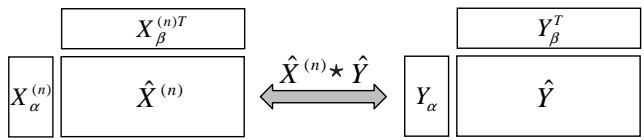}
 \caption{Approximated biometric images $\hat{X}^{(n)}$ and $\hat{Y}$ and the cross-correlation $\hat{X}^{(n)} \star \hat{Y}$. $\hat{X}^{(n)} \star \hat{Y}$ can be computed via a small number of 1D inverse NTTs.}
 \label{fig:MF_corr}
\end{center}
\end{figure}

The proposed indexing scheme is based on our findings that 
the number of the 1D inverse NTTs in CIRF can be significantly reduced 
by approximating biometric images by \textit{low-rank matrices}. 
%
Specifically, 
the proposed scheme approximates $X^{(n)}$ and $Y$ 
by low-rank matrices $\hat{X}^{(n)}$ and $\hat{Y}$, respectively, 
and factorizes 
each of them 
into small matrices 
using matrix factorization 
\cite{Cichocki_Wiley09,Ravanbakhsh_ICML16,Zhang_ICDM07}. 
Then it uses 
small matrices computed from $\hat{X}^{(n)}$ and $\hat{Y}$ 
as indexes of $\hat{X}^{(n)}$ and $\hat{Y}$, respectively. 
The proposed scheme performs pattern matching for the indexes 
in an analogous way to CIRF. 
More specifically, it transforms the indexes 
via the 1D NTT and random filtering, and 
computes the cross-correlation $\hat{X}^{(n)} \star \hat{Y}$ 
from the transformed indexes 
(without restoring the original indexes) 
via a small number of 1D inverse NTTs. 
Fig.~\ref{fig:MF_corr} shows $\hat{X}^{(n)}$, $\hat{Y}$, and $\hat{X}^{(n)} \star \hat{Y}$ 
($\hat{X}_\alpha^{(n)}$, $\hat{X}_\beta^{(n)}$, $\hat{Y}_\alpha$, and $\hat{Y}_\beta$ 
are small matrices, which are described in detail in Section~\ref{sub:algorithm}). 

It should be noted that 
since $\hat{X}^{(n)}$ and $\hat{Y}$ approximate $X^{(n)}$ and $Y$, respectively,  
$\hat{X}^{(n)} \star \hat{Y}$ also approximates $X^{(n)} \star Y$. 
Let $\hat{s}: \mathbf{X} \times \mathbf{Y} \rightarrow \mathbb{R}$ be an 
\textit{approximate score function} 
that takes input $X^{(n)}$ and $Y$ and outputs an approximate score 
(distance or similarity) 
$\hat{s}(X^{(n)},Y) \in \mathbb{R}$ between $X^{(n)}$ and $Y$. 
The proposed scheme computes an approximate score $\hat{s}(X^{(n)},Y)$ 
using 
$\hat{X}^{(n)} \star \hat{Y}$, 
and sorts $N$ transformed templates in ascending (or descending) order of 
the approximate distance (or similarity) $\hat{s}(X^{(n)},Y)$. 
Then it computes an exact score $s(X^{(n)},Y)$ according to the sorted order. 
Since 
$\hat{s}(X^{(n)},Y)$ is highly correlated with $s(X^{(n)},Y)$, 
a genuine template can be found in the early stage of exact matching. 

The proposed indexing scheme can be applied to any kind of biometric traits whose score is measured via cross-correlation. 
In this paper, we use CIRF to compute an exact score, 
since it is promising with regard to the properties 
(i), (ii), (iii), and (iv). 


\subsection{Algorithm}
\label{sub:algorithm}

We now describe details of the proposed indexing scheme. 
We regard the $n$-th template $X^{(n)}$ and the query sample $Y$ 
as \textit{rank-k matrices} 
with very small $k$ 
($k \ll \min\{h, w\}$; in our experiments, 
$k=1$ or $2$), 
and approximate them as follows: 
\begin{align}
\hat{X}^{(n)} &= X_\alpha^{(n)} X_\beta^{(n)T} 
= \sum_{i=1}^k x_{\alpha i}^{(n)} x_{\beta i}^{(n)T} \\
\hat{Y} &= Y_\alpha Y_\beta^T 
= \sum_{i=1}^k y_{\alpha i} y_{\beta i}^T
\end{align}
where 
$\hat{X}^{(n)} \in \mathbb{Z}_p^{h \times w}$, 
$\hat{Y} \in \mathbb{Z}_p^{h \times w}$, 
$X_\alpha^{(n)}, Y_\alpha \in \mathbb{Z}_p^{h \times k}$, $X_\beta^{(n)}, Y_\beta \in \mathbb{Z}_p^{w \times k}$, 
$x_{\alpha i}^{(n)}, y_{\alpha i} \in \mathbb{Z}_p^h$, 
and 
$x_{\beta i}^{(n)}, y_{\beta i} \in \mathbb{Z}_p^w$. 

We define $X_{idx}^{(n)}$ $(1 \leq n \leq N)$ and $Y_{idx}$ as follows:
\begin{align}
X_{idx}^{(n)} &= \{X_\alpha^{(n)}, X_\beta^{(n)}\} 
= \{x_{\alpha i}^{(n)}, x_{\beta i}^{(n)} | 1 \leq i \leq k \} \\
Y_{idx} &= \{Y_\alpha, Y_\beta\} 
= \{y_{\alpha i}^{(n)}, y_{\beta i}^{(n)} | 1 \leq i \leq k \}, 
\end{align}
and use them 
as an {\it index} of 
$X^{(n)}$ and $Y$, respectively. 
To compute $X_{idx}^{(n)}$ and $Y_{idx}$ from $X^{(n)}$ and $Y$, 
we use a matrix factorization method 
such as NMF (Non-negative Matrix Factorization) \cite{Cichocki_Wiley09} and BMF 
(Boolean Matrix Factorization) \cite{Ravanbakhsh_ICML16,Zhang_ICDM07}. 

\begin{figure}
\centering
 \includegraphics[width=0.95\linewidth]{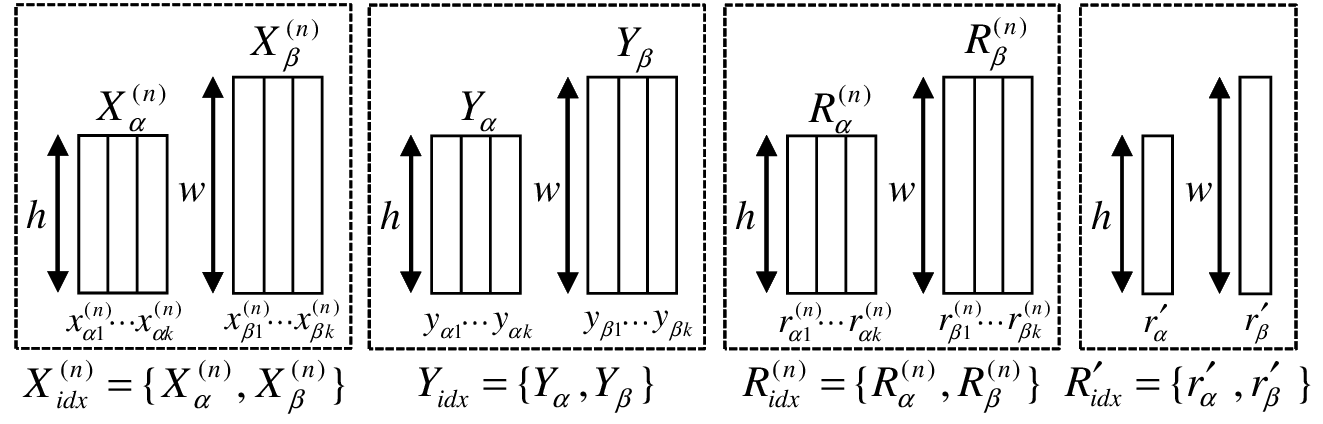}
 \caption{Illustration of $X_{idx}^{(n)}$ (index of $X^{(n)}$), $Y_{idx}$ (index of $Y$), $R_{idx}^{(n)}$ (parameter for $X_{idx}^{(n)}$), and $R'_{idx}$ 
(additional parameter for $X_{idx}^{(1)}$). }
 \label{fig:MF}
\end{figure}

We use 
\begin{align}
R_{idx}^{(n)} &= \{R_\alpha^{(n)}, R_\beta^{(n)}\} 
= \{r_{\alpha i}^{(n)}, r_{\beta i}^{(n)} | 1 \leq i \leq k \} 
\end{align}
($R_\alpha^{(n)} \in (\mathbb{Z}_p^*)^{h \times k}$, $R_\beta^{(n)} \in (\mathbb{Z}_p^*)^{w \times k}$, $r_{\alpha i}^{(n)} \in (\mathbb{Z}_p^*)^h$, $r_{\beta i}^{(n)} \in (\mathbb{Z}_p^*)^w$; 
$r_{\alpha i}^{(n)}$ and $r_{\beta i}^{(n)}$ are random vectors) 
as a parameter (random filter) for the index $X_{idx}^{(n)}$ ($1 \leq n \leq N$). 
For the first index $X_{idx}^{(1)}$, we also use 
\begin{align}
R'_{idx} = \{r'_\alpha, r'_\beta\} 
\end{align}
($r'_\alpha \in (\mathbb{Z}_p^*)^h$, $r'_\beta \in (\mathbb{Z}_p^*)^w$; they are random vectors) as an additional parameter. 
$R'_{idx}$ is required (in addition to $R_{idx}^{(n)}$) 
to compute the cyclic convolution $\hat{X}^{(n)} \star \hat{Y}$, as 
explained later in detail.  
$R_{idx}^{(n)}$ ($1 \leq n \leq N$) and $R'_{idx}$ are stored in a client or a parameter management server \cite{Takahashi_CIBIM11}. 
Fig.~\ref{fig:MF} shows $X_{idx}^{(n)}$, $Y_{idx}$, $R_{idx}^{(n)}$, and $R'_{idx}$. 
In the following, we describe the proposed algorithm in the individual parameter scenario, 
where $R_{idx}^{(n)}$ ($1 \leq n \leq N$) and $R'_{idx}$ are independently and uniformly generated. 
The algorithm below can easily be extended to the common parameter scenario by 
setting $R_{idx}^* = R_{idx}^{(1)} = \cdots = R_{idx}^{(N)}$ and using $R_{idx}^*$ 
as a common parameter 
($R'_{idx}$ is also used in the common parameter scenario in the same way as the individual parameter scenario). 

\begin{figure}
\begin{center}
 \includegraphics[width=0.95\linewidth]{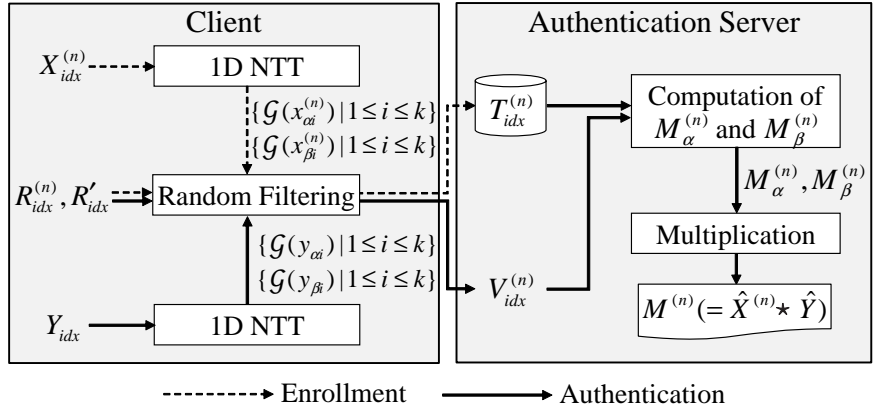}
 \caption{Proposed indexing scheme ($\mathcal{G}$: 1D NTT, $M^{(n)}$ ($= \hat{X}^{(n)} \star \hat{Y}$): approximation of $X^{(n)} \star Y$). }
 \label{fig:proposed}
\end{center}
\end{figure}

Fig.~\ref{fig:proposed} shows the process of the proposed indexing scheme 
in the enrollment/authentication phase. 
In the enrollment phase, the proposed indexing scheme performs the following process: 
\begin{enumerate}
\item Compute transformed indexes $T_{idx}^{(n)}$ $(1 \leq n \leq N)$ from 
$X_{idx}^{(n)}$ $(1 \leq n \leq N)$, $R_{idx}^{(n)}$ $(1 \leq n \leq N)$, 
and $R'_{idx}$ as follows: 
\begin{align}
T_{idx}^{(n)} = 
\begin{cases}
\{t_{\alpha i}^{(1)} , t_{\beta i}^{(1)}, t'_\alpha, t'_\beta | 1 \leq i \leq k \} 	&	(\text{if } n = 1) \\
\{t_{\alpha i}^{(n)} , t_{\beta i}^{(n)} | 1 \leq i \leq k \} 	&	(\text{if } 2 \leq n \leq N),
\label{eq:T_idx_n}
\end{cases}
\end{align}
where 
\begin{align}
t_{\alpha i}^{(n)} &= \mathcal{G}(x_{\alpha i}^{(n)}) \circ r_{\alpha i}^{(n)} ~~ (1 \leq n \leq N, 1 \leq i \leq k) \label{eq:t_alpha_i_n}\\
t_{\beta i}^{(n)} &= \mathcal{G}(x_{\beta i}^{(n)}) \circ r_{\beta i}^{(n)} ~~ (1 \leq n \leq N, 1 \leq i \leq k) \label{eq:t_beta_i_n}\\
t'_\alpha &= \mathcal{G}(x_{\alpha 1}^{(1)}) \circ r'_\alpha \label{eq:t_prime_alpha}\\
t'_\beta &= \mathcal{G}(x_{\beta 1}^{(1)}) \circ r'_\beta \label{eq:t_prime_beta}
\end{align}
and $\mathcal{G}$ denotes the 1D NTT. 
\item Store the transformed indexes $\{T_{idx}^{(n)} | 1 \leq n \leq N\}$ 
in the database of the authentication server 
(along with the transformed templates $\{T^{(n)} | 1 \leq n \leq N\}$). 
\end{enumerate}
In the step 1, we compute 
$t_{\alpha i}^{(n)}$ (resp.~$t_{\beta i}^{(n)}$) in (\ref{eq:t_alpha_i_n}) (resp.~(\ref{eq:t_beta_i_n})) 
by transforming $x_{\alpha i}^{(n)}$ (resp.~($x_{\beta i}^{(n)}$)) via the 1D NTT and filtering 
$r_{\alpha i}^{(n)}$ (resp.~$r_{\beta i}^{(n)}$).
We also compute 
$t'_\alpha$ (resp.~$t'_\beta$) in (\ref{eq:t_prime_alpha}) (resp.~(\ref{eq:t_prime_beta})) 
by filtering an additional parameter $r'_\alpha$ (resp.~$r'_\beta$). 
The size of $\{T_{idx}^{(n)} | 1 \leq n \leq N\}$ is $(h+w)(kn + 1)$ pixels in total. 

In the authentication 
phase, 
the proposed scheme performs the following process: 
\begin{enumerate}
\item Compute transformed indexes $V_{idx}^{(n)}$ $(1 \leq n \leq N)$ from 
$Y_{idx}$, $R_{idx}^{(n)}$ $(1 \leq n \leq N)$, 
and $R'_{idx}$ as follows: 
\begin{align}
&V_{idx}^{(n)} \nonumber \\
&= 
\begin{cases}
\{v_{\alpha i}^{(1)}, v_{\beta i}^{(1)}, v'_{\alpha j}, v'_{\beta j} | 1 \leq i \leq k, 2 \leq j \leq k \} 	&	\hspace{-2mm} (\text{if } n = 1) \\
\{v_{\alpha i}^{(n)}, v_{\beta i}^{(n)} | 1 \leq i \leq k \} 	&	\hspace{-2mm} (\text{if } 2 \leq n \leq N),
\label{eq:V_idx_n}
\end{cases}
\end{align}
where 
\begin{align}
v_{\alpha i}^{(n)} &= \mathcal{G}(\overleftarrow{y_{\alpha i}}) \circ (r_{\alpha i}^{(n)})^{-1} ~~ (1 \leq n \leq N, 1 \leq i \leq k) \label{eq:v_alpha_i_n}\\
v_{\beta i}^{(n)} &= \mathcal{G}(\overleftarrow{y_{\beta i}}) \circ (r_{\beta i}^{(n)})^{-1} ~~ (1 \leq n \leq N, 1 \leq i \leq k) \label{eq:v_beta_i_n}\\
v'_{\alpha j} &= \mathcal{G}(\overleftarrow{y_{\alpha j}}) \circ (r'_\alpha)^{-1} ~~ (2 \leq j \leq k) \label{eq:v_prime_alpha_j}\\
v'_{\beta j} &= \mathcal{G}(\overleftarrow{y_{\alpha j}}) \circ (r'_\beta)^{-1} ~~ (2 \leq j \leq k) \label{eq:v_prime_beta_j}
\end{align}
and $\overleftarrow{y}$ is a flipped vector of $y$ 
(i.e., $\overleftarrow{y_{\alpha i}}[j] = y_{\alpha i}[h - j - 1]$, 
$\overleftarrow{y_{\beta i}}[j] = y_{\beta i}[w - j - 1]$). 
\item Compute matrices $M_\alpha^{(n)} \in \mathbb{Z}_p^{h \times k^2}$ 
and $M_\beta^{(n)} \in \mathbb{Z}_p^{w \times k^2}$ ($1 \leq n \leq N$), which are given by 
\begin{align}
M_\alpha^{(n)} &= [\mathcal{G}^{-1}(\mathcal{G}(x_{\alpha i}^{(n)}) \circ \mathcal{G}(\overleftarrow{y_{\alpha j}}))]_{1 \leq i,j \leq k} \label{eq:M_alpha_n} \\
M_\beta^{(n)} &= [\mathcal{G}^{-1}(\mathcal{G}(x_{\beta i}^{(n)}) \circ \mathcal{G}(\overleftarrow{y_{\beta j}}))]_{1 \leq i,j \leq k}, \label{eq:M_beta_n}
\end{align}
where $\mathcal{G}^{-1}$ is the 1D inverse NTT and 
$[a]_{1 \leq i,j \leq k}$ is a matrix of $k^2$ columns, 
whose ($(i-1)k+j$)-th column 
is given by $a$. 
We can compute 
$M_\alpha^{(n)}$ and $M_\beta^{(n)}$ 
from $T_{idx}^{(n)}$ and $V_{idx}^{(n)}$ 
via 
minimum spanning trees, 
as described 
in Section~\ref{sub:computation_M}. 
\item Compute the following matrix $M^{(n)} \in \mathbb{Z}_p^{h \times w}$ ($1 \leq n \leq N$): 
\begin{align}
M^{(n)} = M_\alpha^{(n)} M_\beta^{(n)T}.
\label{M_n}
\end{align}


As we prove later, $M^{(n)} = \hat{X}^{(n)} \star \hat{Y}$ 
(Section~\ref{sub:analysis}, \textbf{Theorem 1}). 
In other words, $M^{(n)}$ approximates the cross-correlation $X^{(n)} \star Y$, 
and therefore 
we can compute 
an {\it approximate score} $\hat{s}(X^{(n)}, Y)$ 
based on 
$M^{(n)}$. 

Note that 
the number of the 1D inverse NTTs 
necessary to compute $M_\alpha^{(n)}$ in (\ref{eq:M_alpha_n}) is $k^2$. 
Similarly, 
the number of the 1D inverse NTTs 
necessary to compute $M_\beta^{(n)}$ in (\ref{eq:M_beta_n}) is $k^2$.  
Thus, \textit{the total number of the 1D inverse NTTs 
necessary to compute $M^{(n)}$ is $2k^2$}, 
which is much smaller than that in CIRF 
(i.e., $2k^2 \ll h+w$) when $k$ is very small.
\item Sort $N$ transformed templates in ascending (or descending) order of 
the 
approximate distance (or similarity) $\hat{s}(X^{(n)}, Y)$. 
\end{enumerate}
Then, the proposed scheme computes an 
exact 
score $s(X^{(n)},Y)$ 
based on $X^{(n)} \star Y$ 
according to the sorted order. 
If 
the exact distance (or similarity) $s(X^{(n)},Y)$ falls below (or exceeds) a threshold (i.e., if a genuine template is found), 
it identifies the user as the corresponding enrollee and 
terminates the identification process. 
If a genuine template 
is not found after matching all the templates, it rejects the user. 

\subsection{Computation of $M_\alpha^{(n)}$ and $M_\beta^{(n)}$ via minimum spanning trees}
\label{sub:computation_M}
We now explain how to compute 
$M_\alpha^{(n)}$ and $M_\beta^{(n)}$ ($1 \leq n \leq N$) 
in (\ref{eq:M_alpha_n}) and (\ref{eq:M_beta_n}) 
using 
the transformed indexes 
$T_{idx}^{(n)}$ and $V_{idx}^{(n)}$ ($1 \leq n \leq N$), 
which are given by 
(\ref{eq:T_idx_n}) and (\ref{eq:V_idx_n}), respectively.  
We begin by explaining how to compute $M_\alpha^{(n)}$ ($1 \leq n \leq N$) in detail 
(since $M_\beta^{(n)}$ can be computed in the same way as $M_\alpha^{(n)}$). 

\begin{figure}
\begin{center}
\includegraphics[width=0.8\linewidth]{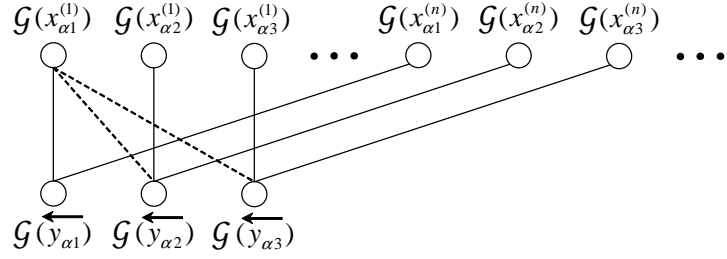}
 \caption{Minimum spanning tree formed from the equations (\ref{eq:G_x_i_n_G_y_i}) and (\ref{eq:G_x_1_1_G_y_i}) ($k=3$). Solid (resp.~dash) lines represent the Hadamard products in  (\ref{eq:G_x_i_n_G_y_i}) (resp.~(\ref{eq:G_x_1_1_G_y_i})).}
 \label{fig:min_span_tree}
\end{center}
\end{figure}

First, 
we compute 
\begin{align}
t_{\alpha i}^{(n)} \circ v_{\alpha i}^{(n)} &= \mathcal{G}(x_{\alpha i}^{(n)}) \circ \mathcal{G}(\overleftarrow{y_{\alpha i}}) ~~ (1 \leq n \leq N, 1 \leq i \leq k) \label{eq:G_x_i_n_G_y_i} \\
t'_{\alpha} \circ v'_{\alpha j} &= \mathcal{G}(x_{\alpha 1}^{(1)}) \circ \mathcal{G}(\overleftarrow{y_{\alpha j}}) ~~ (2 \leq j \leq k) \label{eq:G_x_1_1_G_y_i}
\end{align}
using 
$T_{idx}^{(n)}$ and $V_{idx}^{(n)}$ 
(see 
(\ref{eq:t_alpha_i_n}), 
(\ref{eq:t_prime_alpha}), 
(\ref{eq:v_alpha_i_n}), and 
(\ref{eq:v_prime_alpha_j})). 
It is important to note that 
(\ref{eq:G_x_i_n_G_y_i}) and (\ref{eq:G_x_1_1_G_y_i}) 
form a {\it minimum spanning tree} \cite{algorithms}, 
whose vertices represent 
$\mathcal{G}(x_{\alpha i}^{(n)})$ and $\mathcal{G}(\overleftarrow{y_{\alpha i}})$ ($1 \leq n \leq N, 1 \leq i \leq k$), and 
whose edges represent their Hadamard products. 
Fig.~\ref{fig:min_span_tree} shows the minimum spanning tree formed from (\ref{eq:G_x_i_n_G_y_i}) and (\ref{eq:G_x_1_1_G_y_i}) in the case 
where 
$k=3$. 

We compute 
\begin{align}
\{\mathcal{G}(x_{\alpha i}^{(n)}) \circ \mathcal{G}(\overleftarrow{y_{\alpha j}})) | 1 \leq n \leq N, 1 \leq i,j \leq k\} 
\label{eq:G_x_i_n_G_y_j}
\end{align}
using 
the minimum spanning tree. 
Specifically, we compute the Hadamard product of $\mathcal{G}(x_{\alpha i}^{(n)})$ and $\mathcal{G}(\overleftarrow{y_{\alpha j}})$  ($1 \leq n \leq N, 1 \leq i, j \leq k$)  
other than (\ref{eq:G_x_i_n_G_y_i}) and (\ref{eq:G_x_1_1_G_y_i}) 
by tracking the path from $\mathcal{G}(x_{\alpha i}^{(n)})$ to $\mathcal{G}(\overleftarrow{y_{\alpha j}})$ 
via $\mathcal{G}(\overleftarrow{y_{\alpha i}})$ and $\mathcal{G}(x_{\alpha 1}^{(1)})$. 
In other words, 
we compute $\mathcal{G}(x_{\alpha i}^{(n)}) \circ \mathcal{G}(\overleftarrow{y_{\alpha j}})$ 
using (\ref{eq:G_x_i_n_G_y_i}) and (\ref{eq:G_x_1_1_G_y_i}) as follows: 
\begin{align}
&\mathcal{G}(x_{\alpha i}^{(n)}) \circ \mathcal{G}(\overleftarrow{y_{\alpha j}}) \nonumber \\
&= 
(\mathcal{G}(x_{\alpha i}^{(n)}) \circ \mathcal{G}(\overleftarrow{y_{\alpha i}})) \circ 
(\mathcal{G}(x_{\alpha 1}^{(1)}) \circ \mathcal{G}(\overleftarrow{y_{\alpha i}}))^{-1} \circ 
(\mathcal{G}(x_{\alpha 1}^{(1)}) \circ \mathcal{G}(\overleftarrow{y_{\alpha j}})).
\label{eq:track_min_span_tree}
\end{align}
For example, 
we compute 
$\mathcal{G}(x_{\alpha 2}^{(n)}) \circ \mathcal{G}(\overleftarrow{y_{\alpha 3}})$ in Fig.~\ref{fig:min_span_tree} 
by tracking the path from $\mathcal{G}(x_{\alpha 2}^{(n)})$ to $\mathcal{G}(\overleftarrow{y_{\alpha 3}})$ 
via $\mathcal{G}(\overleftarrow{y_{\alpha 2}})$ and $\mathcal{G}(x_{\alpha 1}^{(1)})$: 
$\mathcal{G}(x_{\alpha 2}^{(n)}) \circ \mathcal{G}(\overleftarrow{y_{\alpha 3}}) = 
(\mathcal{G}(x_{\alpha 2}^{(n)}) \circ \mathcal{G}(\overleftarrow{y_{\alpha 2}})) \circ 
(\mathcal{G}(x_{\alpha 1}^{(1)}) \circ \mathcal{G}(\overleftarrow{y_{\alpha 2}}))^{-1} \circ 
(\mathcal{G}(x_{\alpha 1}^{(1)}) \circ \mathcal{G}(\overleftarrow{y_{\alpha 3}}))$. 
After computing (\ref{eq:G_x_i_n_G_y_j}) in this 
manner, 
we compute 
$M_\alpha^{(n)}$ ($1 \leq n \leq N$) in (\ref{eq:M_alpha_n}) 
via the 1D inverse NTTs. 

Similarly, we compute $M_\beta^{(n)}$ ($1 \leq n \leq N$) in (\ref{eq:M_beta_n}) 
using 
the minimum spanning tree. 
Specifically, we compute 
\begin{align}
t_{\beta i}^{(n)} \circ v_{\beta i}^{(n)} &= \mathcal{G}(x_{\beta i}^{(n)}) \circ \mathcal{G}(\overleftarrow{y_{\beta i}}) ~~ (1 \leq n \leq N, 1 \leq i \leq k) \label{eq:min_span_tree_beta1}\\
t'_{\beta} \circ v'_{\beta j} &= \mathcal{G}(x_{\beta 1}^{(1)}) \circ \mathcal{G}(\overleftarrow{y_{\beta j}}) ~~ (2 \leq j \leq k) \label{eq:min_span_tree_beta2}
\end{align}
using 
$T_{idx}^{(n)}$ and $V_{idx}^{(n)}$ 
(see (\ref{eq:t_beta_i_n}), (\ref{eq:t_prime_beta}), (\ref{eq:v_beta_i_n}), and (\ref{eq:v_prime_beta_j})), 
and compute 
\begin{align}
\{\mathcal{G}(x_{\beta i}^{(n)}) \circ \mathcal{G}(\overleftarrow{y_{\beta j}})) | 1 \leq n \leq N, 1 \leq i,j \leq k\} 
\label{eq:G_x_i_n_G_y_j_beta}
\end{align}
using 
a minimum spanning tree formed from 
(\ref{eq:min_span_tree_beta1}) and (\ref{eq:min_span_tree_beta2}). 
After computing (\ref{eq:G_x_i_n_G_y_j_beta}), 
we compute $M_\beta^{(n)}$ ($1 \leq n \leq N$) in (\ref{eq:M_beta_n}) via the 1D inverse NTTs. 

The total number of the Hadamard products 
necessary 
to compute 
(\ref{eq:G_x_i_n_G_y_j}) and (\ref{eq:G_x_i_n_G_y_j_beta}) 
is 
quadratic in the number of $k$. 
However, in our experiments in Section~\ref{sec:exp}, 
we confirmed that the time to compute all of the Hadamard products 
was much 
less 
than the time to compute all of the 1D inverse NTTs.

\subsection{Theoretical properties}
\label{sub:analysis}
We show some theoretical properties of the proposed indexing scheme. 
We begin with the following theorem: 

\begin{theorem}
\begin{align}
M^{(n)} = \hat{X}^{(n)} \star \hat{Y} ~~~~ (1 \leq n \leq N). 
\end{align}
\end{theorem}
The proof is given in \ref{app:efficiency}. 
{\bf Theorem 1} means that 
$M^{(n)}$ approximates $X^{(n)} \star Y$. 
By sorting transformed templates in ascending (or descending) order of 
an approximate distance (or similarity) $\hat{s}(X^{(n)}, Y)$ 
based on $M^{(n)}$, 
we can find a 
genuine template 
in the early 
stage 
of exact matching. 

We then consider the security of the proposed indexing scheme against \textbf{Attackers A}, \textbf{B}, and  \textbf{C} in Section~\ref{sub:desirable_cancelable}. 
We assume that these attackers obtain transformed indexes in addition to 
transformed features, as described in Section~\ref{sub:desirable_cancelable}. 
Let 
$T^{all} = \{T^{(n)} | 1 \leq n \leq N\}$ and 
$T_{idx}^{all} = \{T_{idx}^{(n)} | 1 \leq n \leq N\}$ 
be sets of $N$ transformed templates and $N$ transformed indexes, respectively. 
\textbf{Attacker A} obtains $T^{(n)}$ and $T_{idx}^{(n)}$. 
whereas \textbf{Attacker B} obtains $T^{all}$ and $T_{idx}^{all}$. 
Let further 
$V^{all} = \{V^{(n)} | 1 \leq n \leq N\}$ and 
$V_{idx}^{all} = \{V_{idx}^{(n)} | 1 \leq n \leq N\}$ 
be sets of $N$ transformed query samples and $N$ transformed indexes, respectively. 
\textbf{Attacker C} obtains $T^{all}$, $T_{idx}^{all}$, $V^{all}$, and $V_{idx}^{all}$. 
Table~\ref{tab:attacker_info} summarizes the information available to each attacher.

Let $\mathbf{X}_{idx}$, $\mathbf{T}_{idx}$, 
$\mathbf{T}^{all}$, and $\mathbf{T}_{idx}^{all}$ 
be spaces of 
$X_{idx}^{(n)}$, $T^{(n)}$, 
$T^{all}$, and $T_{idx}^{all}$, respectively. 
We firstly consider 
\textbf{Attacker A}: 
\begin{theorem}
For any 
$X^{(n)} \in \mathbf{X}$, $X_{idx}^{(n)} \in \mathbf{X}_{idx}$, 
$T^{(n)} \in \mathbf{T}$, and $T_{idx}^{(n)} \in \mathbf{T}_{idx}$, 
\begin{align}
\Pr(X^{(n)} | T^{(n)}, T_{idx}^{(n)}) &= \Pr(X^{(n)}) ~~~~ (1 \leq n \leq N) \label{eq:threorem2_X}\\
\Pr(X_{idx}^{(n)} | T^{(n)}, T_{idx}^{(n)}) &= \Pr(X_{idx}^{(n)}) ~~~~ (1 \leq n \leq N) \label{eq:threorem2_X_idx}. 
\end{align}
\end{theorem}
The proof is given in \ref{app:security}. 
{\bf Theorem 2} means that the $n$-th transformed template $T^{(n)}$ and the $n$-th transformed index $T_{idx}^{(n)}$ 
leak no information 
about 
the original transformed template $X^{(n)}$ and the original index $X_{idx}^{(n)}$. 
In other words, the proposed scheme has perfect secrecy against \textbf{Attacker A}. 
Note that this theorem holds for 
both the individual parameter scenario and the common parameter scenario. 

We secondly consider \textbf{Attacker B} in the individual parameter scenario. 
Since parameters $R^{(n)}$  ($1 \leq n \leq N$), $R_{idx}^{(n)}$ ($1 \leq n \leq N$), and $R'_{idx}$ are independent 
in the individual parameter scenario, 
the following theorem is immediately derived from {\bf Threorem 2}: 
\begin{theorem}
In the individual parameter scenario, 
for any 
$X^{(n)} \in \mathbf{X}$, $X_{idx}^{(n)} \in \mathbf{X}_{idx}$, $T^{all} \in \mathbf{T}^{all}$, and $T_{idx}^{all} \in \mathbf{T}^{all}$, 
\begin{align}
\Pr(X^{(n)} | T^{all}, T_{idx}^{all}) &= \Pr(X^{(n)}) ~~~~ (1 \leq n \leq N) \label{eq:threorem3_X}\\
\Pr(X_{idx}^{(n)} | T^{all}, T_{idx}^{all}) &= \Pr(X_{idx}^{(n)}) ~~~~ (1 \leq n \leq N) \label{eq:threorem3_X_idx}. 
\end{align}
\end{theorem}
{\bf Theorem 3} means that the proposed scheme has perfect secrecy against \textbf{Attacker B}  in the individual parameter scenario. 

We thirdly consider \textbf{Attacker C} in the individual parameter scenario. 
In this case, 
$T^{(n)} = \mathcal{F}(X^{(n)}) \circ R^{(n)}$ and $V^{(n)} = \mathcal{F}(\overleftarrow{Y}) \circ (R^{(n)})^{-1}$. 
In addition, $T_{idx}^{(n)}$ and $V_{idx}^{(n)}$ are decomposed into  (\ref{eq:t_alpha_i_n})-(\ref{eq:t_prime_beta}) and (\ref{eq:v_alpha_i_n})-(\ref{eq:v_prime_beta_j}), respectively.  
Thus, the information available to \textbf{Attacker C} is the following system of equations: 
\begin{numcases}{}
T^{(n)} = \mathcal{F}(X^{(n)}) \circ R^{(n)} ~~ (1 \leq n \leq N) \label{eq:syseq1}\\
V^{(n)} = \mathcal{F}(\overleftarrow{Y}) \circ (R^{(n)})^{-1} ~~ (1 \leq n \leq N) \label{eq:syseq2}\\
\text{equations } (\ref{eq:t_alpha_i_n})\text{-}(\ref{eq:t_prime_beta}) \label{eq:syseq3}\\
\text{equations } (\ref{eq:v_alpha_i_n})\text{-}(\ref{eq:v_prime_beta_j}), \label{eq:syseq4}
\end{numcases}
which is equivalent to the following system of equations: 
\begin{numcases}{}
T^{(n)} = \mathcal{F}(X^{(n)}) \circ R^{(n)} ~~ (1 \leq n \leq N) \label{eq:syseq5}\\
\mathcal{F}^{(-1)} (T^{(n)} \circ V^{(n)}) = X^{(n)} \star Y ~~ (1 \leq n \leq N) \label{eq:syseq6}\\
\text{equations } (\ref{eq:t_alpha_i_n})\text{-}(\ref{eq:t_prime_beta}) \label{eq:syseq7}\\
\text{equations } (\ref{eq:G_x_i_n_G_y_i}), (\ref{eq:G_x_1_1_G_y_i}), (\ref{eq:min_span_tree_beta1}), 
\text{and } (\ref{eq:min_span_tree_beta2}) \label{eq:syseq8}
\end{numcases}
((\ref{eq:syseq6}) is obtained by multiplying (\ref{eq:syseq1}) by (\ref{eq:syseq2}) and performing the 2D inverse NTT; (\ref{eq:syseq8}) is obtained by multiplying (\ref{eq:syseq3}) by (\ref{eq:syseq4})). 
From \textbf{Theorem 3}, 
(\ref{eq:syseq5}) and (\ref{eq:syseq7}) provide 
no information about $X^{(n)}$ and $X_{idx}^{(n)}$ ($1 \leq n \leq N$). 
Thus, the only information available to \textbf{Attacker C} is 
(\ref{eq:syseq6}) and (\ref{eq:syseq8}), 
which are necessary to compute scores $s(X^{(n)}, Y)$ ($1 \leq n \leq N$) 
and approximate scores $\hat{s}(X^{(n)}, Y)$ ($1 \leq n \leq N$), respectively. 

We now consider an attack that tries to recover $X^{(n)}$, $Y$, $X_{idx}^{(n)}$, and $Y_{idx}$ 
by solving (\ref{eq:syseq6}) and (\ref{eq:syseq8}). 
(\ref{eq:syseq6}) is a system of quadratic simultaneous equations with 
$(N+1)hw$ unknown variables (i.e., $X^{(1)}, \cdots, X^{(N)}$, and $Y$) and $Nhw$ equations. 
(\ref{eq:syseq8}) is a system of quadratic simultaneous equations with 
$(Nk+k)(h+w)$ unknown variables (i.e., $X_{idx}^{(1)}, \cdots, X_{idx}^{(N)}$, and $Y_{idx}$) and 
$(Nk+k-1)(h+w)$ equations. 
Thus, the number of unknown variables is larger than the number of equations in both (\ref{eq:syseq6}) and (\ref{eq:syseq8}), and it is hard to recover $X^{(n)}$, $Y$, $X_{idx}^{(n)}$, and $Y_{idx}$ from these 
equations. 
\begin{table}[t]
\caption{Information available to \textbf{Attackers A}, \textbf{B}, and  \textbf{C}.}
\centering
\hbox to\hsize{\hfil
\begin{tabular}{c|c|c}\hline
\hline
\textbf{Attacker A}		&	\textbf{Attacker B}	&	\textbf{Attacker C}\\
\hline
$T^{(n)}$, $T_{idx}^{(n)}$	&	$T^{all}$, $T_{idx}^{all}$	&	$T^{all}$, $T_{idx}^{all}$, $V^{all}$, $V_{idx}^{all}$\\
\hline
\end{tabular}
\hfil}
\label{tab:attacker_info}
\caption{Security of the proposed indexing scheme (\textbf{A}: \textbf{Attacker A}, 
\textbf{B}: \textbf{Attacker B}, \textbf{C}: \textbf{Attacker C}, 
\#unknown: the number of unknown variables). }
\centering
\hbox to\hsize{\hfil
\begin{tabular}{c||c|c}\hline
\hline
							&	individual parameter						&	common parameter\\
\hline
\textbf{A}		&	perfect secrecy								&	perfect secrecy\\
\textbf{B}		&	perfect secrecy								&	\#unknown $>$ \#equations\\
\textbf{C}		&	\#unknown $>$ \#equations		&	\#unknown $>$ \#equations\\
\hline
\end{tabular}
\hfil}
\label{tab:security}
\end{table}

We also consider the security of the proposed scheme 
against \textbf{Attackers B} and \textbf{C} in the common parameter scenario 
where $R^* = R^{(1)} = \cdots = R^{(N)}$ and $R_{idx}^* = R_{idx}^{(1)} = \cdots = R_{idx}^{(N)}$. 
In this case, the system of equations (\ref{eq:syseq1}) and (\ref{eq:syseq2}) are 
quadratic simultaneous equations with 
$(N+2)hw$ unknown variables (i.e., $X^{(1)}, \cdots, X^{(N)}$, $Y$, and $R^*$) 
and $(N+1)hw$ equations. 
Similarly, 
the system of equations (\ref{eq:syseq3}) and (\ref{eq:syseq4}) 
are quadratic simultaneous equations with 
$(Nk+2k+1)(h+w)$ unknown variables 
(i.e., $X_{idx}^{(1)}, \cdots, X_{idx}^{(N)}$, $Y_{idx}$, $R_{idx}^*$, and $R'_{idx}$) 
and $(Nk+2k)(h+w)$ equations. 
Thus, the number of unknown variables is also larger than the number of equations in this case. 
Therefore, it is hard for \textbf{Attacker C} to recover $X^{(n)}$, $Y$, $X_{idx}^{(n)}$, and $Y_{idx}$ 
from these equations. 
Since \textbf{Attacker B} does not obtain the equations (\ref{eq:syseq2}) and (\ref{eq:syseq4}), 
it is harder for her to recover the original data. 

Table~\ref{tab:security} summarizes the security of the proposed scheme. 
The proposed scheme also has the diversity and the revocability in the same way as CIRF described in Section~\ref{sub:algorithm_CIRF} 
(since we can prove them in the same way as CIRF, we omit the proof). 

We finally discuss the communication cost. 
Assume that we use the cross-correlations $X^{(n)} \star Y$ and $\hat{X}^{(n)} \star \hat{Y}$ as an exact score $s(X^{(n)}, Y)$ and an approximate score $\hat{s}(X^{(n)}, Y)$, respectively. 
In the individual parameter scenario, 
the client (or the parameter management server) 
needs 
to send 
$N$ transformed query samples $V^{all} = \{V^{(n)} | 1 \leq n \leq N\}$ and 
$N$ transformed indexes $V_{idx}^{all} = \{V_{idx}^{(n)} | 1 \leq n \leq N\}$, 
whose size is $Nhw + (Nk+k-1)(h+w)$ pixels in total. 
In the common parameter scenario, 
the size is reduced to $hw + (2k-1)(h+w)$ pixels. 

For example, 
if 
each pixel is represented as a short integer, 
$h=32$, $w=64$, $k=2$ (as in our experiments), and 
$N=32000$, 
then 
the total sizes in the individual parameter scenario and the common parameter scenario 
are $143$ megabytes and $4672$ bytes, respectively. 
If we can use the 100 Gigabit Ethernet private line, 
the communication cost does not matter even in the individual parameter scenario. 
The 400 Gigabit Ethernet will also be available in the near future.  
In such cases, we should use an individual parameter, 
since it has perfect secrecy against \textbf{Attacker B}. 
If we cannot use such a high-speed Ethernet and 
the communication cost is a major problem in the individual parameter scenario, 
we should use a common parameter. 

\section{Experimental evaluation}
\label{sec:exp}
\subsection{Experimental set-up}
\label{sub:setup}
We evaluated the proposed indexing scheme using the finger-vein dataset in \cite{Yanagawa_BIC07}, which 
includes 
six fingers (index fingers, middle fingers, and ring fingers of both hands) from $505$ subjects. 
We used this dataset because it 
includes 
more subjects than other finger-vein datasets  \cite{HongKong,Ton_ICB13,SDUMLA-HMT}. 
To further increase the number of subjects, we assumed that index, middle, and ring fingers are presented by different users. 
In other words, we assumed that the dataset in \cite{Yanagawa_BIC07} 
includes 
two fingers (left finger and right finger) from 
each of 
$1515$ subjects. 
We used two images per finger (one 
for enrollment and the other 
for authentication), and 
excluded $32$ subjects whose fingers were not appropriately captured. 
In total, we used 
two fingers (left finger and right finger) from $1483$ ($= 1515 - 32$) subjects. 

We extracted a finger-vein pattern 
from each finger-vein image using the feature extraction method in \cite{Miura_MVA04}, and transformed it into a binary image ($h=32, w=64$; each pixel takes $1$ (vein) or $0$ (background)). 
We set 
maximum allowable shift lengths ($\Delta i_{max}$, $\Delta j_{max}$) 
in computing $X^{(n)} \star Y$ and $M^{(n)}$ $(= \hat{X}^{(n)} \star \hat{Y})$ 
as ($\Delta i_{max}$, $\Delta j_{max}$) = ($6, 12$) and ($2, 4$), respectively 
(we confirmed that these values provided high accuracy). 
We then set zero values for the uppermost $\Delta i_{max}$ pixels, lowermost $\Delta i_{max}$ pixels, 
leftmost $\Delta j_{max}$ pixels, and rightmost $\Delta j_{max}$ pixels of each enrolled image (e.g., zero-padding) to use cyclic cross-correlation. 
As an exact score $s(X^{(n)}, Y)$, 
we computed the minimum of the Hamming distances of overlapped binary images 
via CIRF 
(see \ref{app:Hamming} for how to compute the minimum Hamming distance via CIRF). 
We used (not the cross-correlation $X^{(n)} \star Y$ but) 
the minimum Hamming distance as an exact score, because 
it provided higher identification accuracy than $X^{(n)} \star Y$. 
Regarding $p$, $\alpha$, and $\beta$ in (\ref{eq:F_X_u_v}), 
we set 
$p=8641$, $\alpha=40$, and $\beta=948$, 
respectively.

We assumed that all templates of $1483$ users are enrolled in the authentication server 
(the number of templates is $N=2966$). 
Then we 
performed an experiment, 
where 
each user inputs left and right fingers and 
the system identifies the user. 
It should be noted 
here that 
FAR in identification (the error rate that an non-enrollee is accepted as an enrollee) 
increases 
as the number of enrollees increases \cite{guide}. 
To achieve 
high accuracy, 
we integrated, for each enrollee, two 
exact scores 
from left and right fingers (i.e., score level fusion 
\cite{guide}). 
As a fusion scheme, we used a sum rule, which sums up the two scores, 
since this rule is equivalent to logistic regression \cite{Murakami_TIFS14,Poh12} 
using the same regression coefficients for the two fingers 
(the effectiveness of logistic regression has been shown in score level fusion \cite{Murakami_TIFS14,Poh12}). 
We evaluated  EER (Equal Error Rate; the operating point where FAR equals to FRR \cite{guide}) in the case 
where 
the system identifies the user by computing all 
$2966$ exact scores (i.e., the system does not use an indexing scheme) 
using $1483 \times 1483$ integrated scores. 
The result was EER $=2.0 \times 10^{-3}$. 

Using the proposed indexing scheme described in Section~\ref{sec:cancelable_index}, 
we computed two approximate scores for each enrollee. 
Specifically, we computed 
$M^{(n)}$ ($= \hat{X}^{(n)} \star \hat{Y}$) in (\ref{M_n}), 
and directly 
used it as an approximate score $\hat{s}(X^{(n)},Y)$; 
i.e., $\hat{s}(X^{(n)},Y) = M^{(n)}$. 
Then we 
integrated the two approximate scores using the sum rule, 
and sorted $1483$ enrollees by the integrated approximate scores. 
Here we used BMF in \cite{Zhang_ICDM07} as a factorization method, and 
set the rank $k$ to $k=1$ or $2$. 
Then we computed two exact scores for each of the enrollees according to the sorted order, 
and integrated the two exact scores using the sum rule. 
When the integrated exact score (i.e., the sum of the Hamming distances) fell below a threshold, 
we identified the user as the corresponding enrollee and terminated the identification process. 

We also compared the proposed indexing scheme with an existing indexing scheme. 
Specifically, we focused on the fact that one of the most popular indexing schemes was based on LSH (Locality Sensitive Hashing) \cite{Cappeli_TPAMI11,Kuzu_ICDE12,Shuai_ICPR08,Tang_ICPR10}, 
which computes $l$ $k$-bit hashes for each biometric feature as an index. 
The LSH-based indexing scheme for finger-vein identification was also proposed in \cite{Tang_ICPR10}. 
However, since LSH is only applicable for specific distance measures (e.g., Hamming distance, $L_p$ distance), 
it requires the alignment of two images in finger-vein identification. 
Although the study in \cite{Tang_ICPR10} assumed that the image alignment is successfully performed, 
the image alignment is difficult especially in the case of template protection (since the original template is not available). 
Taking this into account, we 
evaluated 
{\it DBH (Distance-based Hashing)} \cite{Athitsos_ICDE08}, which is a variant of LSH that can be applied to arbitrary distance measures. 
DBH can be applied to finger-vein identification without requiring 
image alignment, and significantly outperforms VP-trees, a well-known distance-based indexing method \cite{Athitsos_ICDE08}. 
Therefore, 
we consider DBH is a good candidate for comparison 
(we do not explain the algorithm for DBH in this paper; see \cite{Athitsos_ICDE08} for details). 
We randomly selected $100$ templates (from $2966$ templates) to construct hash functions (in the same way as \cite{Athitsos_ICDE08}), and 
attempted various values for the parameters $k$ and $l$ from $1$ to $1000$. 
Note that DBH does not protect the original index (and therefore cannot be used for biometric identification over networks), unlike the proposed scheme. 
Nonetheless, it is important to evaluate 
DBH, since it shows 
how efficient the proposed scheme is compared to the existing indexing scheme.

\subsection{Experimental results}
\label{sub:results}

\begin{figure}
\begin{center}
 \includegraphics[width=1.0\linewidth]{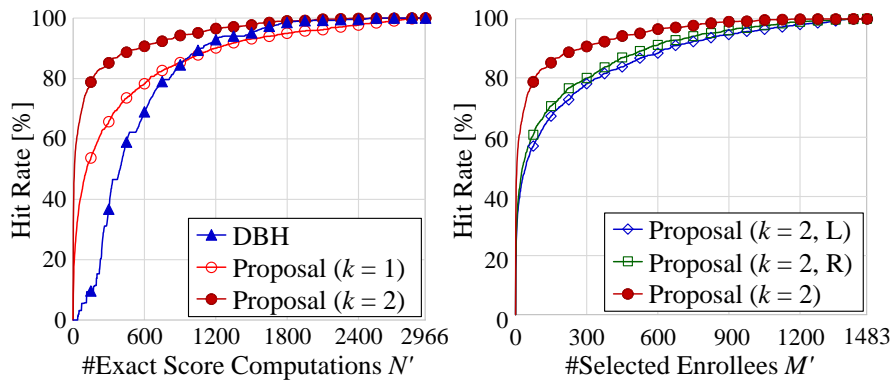}
 \caption{Hit rate [\%]. 
  L and R in the right panel represent the case when a user inputs only a left finger or a right finger, respectively. }
 \label{fig:res_fusion}
\end{center}
\end{figure}

We first 
fixed the number of exact score computations $N'$ ($\leq 2966$), and 
evaluated a {\it hit rate}, a percentage of the cases 
in which 
the first $N'$ templates 
include 
a genuine template. 
The left panel of Fig.~\ref{fig:res_fusion} shows 
the relationship between $N'$ and the hit rate. 
For DBH, we show the best performance obtained by changing $k$ and $l$ for various values from $1$ to $1000$. 
It can be seen that the proposed indexing scheme outperforms DBH. 
We emphasize again that DBH does not protect the original index. 
The proposed scheme protects the original index, as discussed in Section~\ref{sub:analysis}, and provides a higher hit rate than DBH. 

To investigate how the hit rate changes by integrating two approximate scores, 
we also evaluated the hit rate of the proposed scheme 
in the case 
where 
each user inputs only a left or right finger. 
The right panel of Fig.~\ref{fig:res_fusion} shows 
the relationship between the number of selected enrollees $M'$ ($\leq 1483$), whose exact scores have been computed, and the hit rate. 
It can be seen that the hit rate is improved by integrating two approximate scores. 
This is because 
approximate scores are highly correlated with exact distances. 
In other words, the discriminative power of approximate scores can be improved by score level fusion (in the same way as exact scores). 

\begin{table}[t]
\caption{Average Number of exact score computations $\overline{N'}$ 
necessary 
to find a genuine template (i.e., to terminate the identification process) [\%].}
\centering
\hbox to\hsize{\hfil
\begin{tabular}{c||r|r|r}\hline
\hline
							&	DBH		&	Proposal ($k=1$)	&	Proposal ($k=2$)\\	
\hline
$\overline{N'}$ 		&	526.3	&	391.7					&	164.7\\
\hline
\end{tabular}
\hfil}
\label{tab:result}
\end{table}

We then evaluated the average number of exact score computations $\overline{N'}$  
necessary 
to find a genuine template (i.e., to terminate the identification process). 
Table~\ref{tab:result} shows the results. 
In the proposed scheme with rank $k=2$, 
$\overline{N'}$ 
was $164.7$, which is about one-eighteenth of the number of enrollees $N$ ($=2966$). 
We also measured the time to compute one exact score and one approximate score in the proposed scheme with $k=2$ 
on an Intel Xeon CPU E5-2620 v3 ($2.40$ GHz, $6$ cores) with $32$ GB RAM. 
The results were $0.28$ ms and 
$0.015$ 
ms, respectively. 
For example, if the number of templates is 
$N=32000$, it takes about $9$ ($\approx 0.28 \times 10^{-3} \times 32000$) seconds to identify a user in the original CIRF. 
By using the proposed scheme with $k=2$, the identification time can be reduced to about 
one ($\approx 0.015 \times 10^{-3} \times 32000 + \frac{0.28 \times 10^{-3} \times 32000}{18}$) 
second 
on average. 
Note that the proposed scheme 
can significantly reduce the average identification time without affecting 
the identification accuracy (i.e., EER remains to be $2.0 \times 10^{-3}$), since it computes exact scores until a genuine template is found. 

By combining these experimental results with the discussions in Section~\ref{sub:analysis}, 
we 
conclude that the proposed indexing scheme is promising with regard to the properties (i), (ii), (iii), (iv), and (v) in Section~\ref{sub:desirable_cancelable} 
in this dataset. 

However, it should be noted that 
the rank-2 approximation may not be sufficient for other applications. 
For example, Hearn and Reichel \cite{Hearn_ANM14} showed that 
the rank-3 approximation was necessary to 
detect all planes or all spots in cells 
via FFT-based convolution.
The quality of fingerprint images can be poor 
when we use the singular value decomposition (SVD) with rank 1 or 2 \cite{SVD_FINGERPRINT}. 
As future work, we would like to evaluate 
the proposed indexing method using 
other biometric traits such as fingerprint, face, and iris. 
We would also like to improve the proposed indexing scheme so that 
the number of the 1D inverse NTTs is (not $O(k^2)$ but) $O(k)$ to extend its applicability.

\section{Conclusions}
\label{sec:conc}
In this paper, we proposed a cancelable indexing scheme 
based on low-rank approximation of CIRF. 
We proved that 
the transformed index 
leaks no information 
about the original biometric feature and the original index, 
and 
thoroughly discussed the security of the proposed scheme. 
We also 
demonstrated 
that the proposed scheme outperforms DBH \cite{Athitsos_ICDE08}, which is a variant of LSH that can be applied to finger-vein identification, and 
significantly reduces the one-to-many matching time. 

\appendix
\section{Computation of the minimum Hamming distance via CIRF}
\label{app:Hamming}
We explain how to compute the minimum 
of the Hamming distances of overlapped binary images 
(over all values of $\Delta i$ and $\Delta j$) via CIRF. 
Let $\bar{X}$ and $\bar{Y}$ be binary images that flip $0$ and $1$ in each pixel of a template $X$ and a query sample $Y$, respectively. 
At the enrollment phase, 
we randomly and independently generate two parameters $R_1$ and $R_2$, 
and transform $X$ and $\bar{X}$ as follows: 
$T = F_{R_1}(X)$, $\bar{T} = F_{R_2}(\bar{X})$.
Then we store $T$ and $\bar{T}$ in the authentication server. 
At the authentication phase, 
we transform $Y$ and $\bar{Y}$ as follows: 
$V = G_{R_2}(Y)$, $\bar{V} = G_{R_1}(\bar{Y})$. 
Then we send $V$ and $\bar{V}$ to the authentication server. 
We compute the minimum Hamming distance 
$s(X,Y)$ between $X$ and $Y$ 
from $T$, $\bar{T}$, $V$, and $\bar{V}$ via CIRF as follows: 
\begin{align}
s(X,Y) = \min_{\Delta i, \Delta j} ((\bar{X} \star Y)[\Delta i, \Delta j] +  (X \star \bar{Y})[\Delta i, \Delta j]).
\label{eq:d_H}
\end{align}
We need to compute $\bar{X} \star Y$ and $X \star \bar{Y}$ to obtain 
$s(X,Y)$ 
in (\ref{eq:d_H}). 
Therefore, 
the computation of 
$s(X,Y)$ 
requires 
two 2D inverse NTTs 
(i.e., $2(h+w)$ 1D inverse NTTs) in total. 

\section{Proof of Theorem 1}
\label{app:efficiency}
\begin{align}
&M^{(n)} = M_\alpha^{(n)} M_\beta^{(n)T}  \label{eq:thm1_1}\\
&= \sum_{i=1}^k \sum_{j=1}^k 
\mathcal{F}^{-1}( (\mathcal{G}(x_{\alpha i}^{(n)}) \circ \mathcal{G}(\overleftarrow{y_{\alpha j}})) 
(\mathcal{G}(x_{\beta i}^{(n)}) \circ \mathcal{G}(\overleftarrow{y_{\beta j}}))^T)  \label{eq:thm1_3}\\
&= \sum_{i=1}^k \sum_{j=1}^k 
\mathcal{F}^{-1}( (\mathcal{G}(x_{\alpha i}^{(n)}) \circ \mathcal{G}(x_{\beta i}^{(n)})^T) 
(\mathcal{G}(\overleftarrow{y_{\alpha j}}) \circ \mathcal{G}(\overleftarrow{y_{\beta j}}))^T)  \label{eq:thm1_4}\\
&= \mathcal{F}^{-1}(\sum_{i=1}^k (\mathcal{G}(x_{\alpha i}^{(n)}) \circ \mathcal{G}(x_{\beta i}^{(n)})^T) 
(\sum_{j=1}^k \mathcal{G}(\overleftarrow{y_{\alpha j}}) \circ \mathcal{G}(\overleftarrow{y_{\beta j}})^T))  \label{eq:thm1_5}\\
&= \mathcal{F}^{-1}(\sum_{i=1}^k (\mathcal{F}(x_{\alpha i}^{(n)} x_{\beta i}^{(n)T})) 
(\sum_{j=1}^k \mathcal{F}(\overleftarrow{y_{\alpha j}} \overleftarrow{y_{\beta j}}^T)))  \label{eq:thm1_6}\\
&= \mathcal{F}^{-1}(\mathcal{F}(\sum_{i=1}^k x_{\alpha i}^{(n)} x_{\beta i}^{(n)T}) 
(\mathcal{F}(\sum_{j=1}^k \overleftarrow{y_{\alpha j}} \overleftarrow{y_{\beta j}}^T))) 
= \hat{X}^{(n)} \star \hat{Y}  
 \label{eq:thm1_7}
\end{align}
From (\ref{eq:thm1_4}) to (\ref{eq:thm1_5}) and from (\ref{eq:thm1_6}) to (\ref{eq:thm1_7}), we used the linearity of the NTT. 
From (\ref{eq:thm1_1}) to (\ref{eq:thm1_3}) 
and from (\ref{eq:thm1_5}) to (\ref{eq:thm1_6}), we used the separability theorem \cite{handbook_fourier} for the NTT. \qed 

\section{Proof of Theorem 2}
\label{app:security}
Here we provide an outline of the proof of (\ref{eq:threorem2_X}) and (\ref{eq:threorem2_X_idx}) in the case 
where 
$n=1$ 
(we can prove (\ref{eq:threorem2_X}) and (\ref{eq:threorem2_X_idx}) in the case 
where 
$2 \leq n \leq N$ in the same way as the case 
where 
$n=1$). 
We also assume that all elements in 
$\mathcal{G}(x_{\alpha i}^{(1)})$ and $\mathcal{G}(x_{\beta i}^{(1)})$ ($1 \leq i \leq k$) are non-zero 
(we can extend our proof to the case 
in which 
they can 
include 
zero elements in the same way as \cite{Takahashi_IETBio12}). 

We first prove that $\Pr(X_{idx}^{(1)} | T_{idx}^{(1)}) = \Pr(X_{idx}^{(1)})$ for any $X_{idx}^{(1)}$ and any $T_{idx}^{(1)}$. 
This equation holds if there is exactly one parameter pair ($R_{idx}^{(1)}$, $R'_{idx}$) for any $X_{idx}^{(1)}$ and any $T_{idx}^{(1)}$ (\textbf{Lemma 1} in \cite{Takahashi_IEICE11}). 
By (\ref{eq:t_alpha_i_n})\text{-}(\ref{eq:t_prime_beta}), 
there is exactly one such parameter pair ($R_{idx}^{(1)}$, $R'_{idx}$): 
$r_{\alpha i}^{(1)} = t_{\alpha i}^{(1)} / \mathcal{G}(x_{\alpha i}^{(1)})$, 
$r_{\beta i}^{(1)} = t_{\beta i}^{(1)} / \mathcal{G}(x_{\beta i}^{(1)})$, 
$r'_\alpha = t'_\alpha / \mathcal{G}(x_{\alpha 1}^{(1)})$, 
and $r'_\beta = t'_\beta/ \mathcal{G}(x_{\beta 1}^{(1)})$. 
Then, since $T_{idx}^{(1)} \rightarrow X_{idx}^{(1)} \rightarrow X^{(1)}$ (i.e., $T_{idx}^{(1)}$, $X_{idx}^{(1)}$, and $X^{(1)}$ form a Markov chain), 
$\Pr(X^{(1)} | T_{idx}^{(1)}) = \Pr(X^{(1)})$ for any $X^{(1)}$ and any $T_{idx}^{(1)}$ (we can derive this using Bayes' theorem). 

Similarly, $\Pr(X^{(1)} | T^{(1)}) = \Pr(X^{(1)})$ for any $X^{(1)}$ and any $T^{(1)}$ (as described in Section~\ref{sub:algorithm_CIRF}). 
Then, since $T^{(1)} \rightarrow X^{(1)} \rightarrow X_{idx}^{(1)}$, 
$\Pr(X_{idx}^{(1)} | T^{(1)}) = \Pr(X_{idx}^{(1)})$ for any $X_{idx}^{(1)}$ and any $T^{(1)}$. 
Thus, both $T_{idx}^{(1)}$ and $T^{(1)}$ 
leak no information 
about $X_{idx}^{(1)}$ and $X^{(1)}$, and therefore (\ref{eq:threorem2_X}) and (\ref{eq:threorem2_X_idx}) hold 
(we can derive (\ref{eq:threorem2_X}) and (\ref{eq:threorem2_X_idx}) from the above equations using Bayes' theorem). \qed 

\bibliographystyle{plain}
\bibliography{refs_short}

\end{document}